# Inference in Hybrid Networks: Theoretical Limits and Practical Algorithms

**Uri Lerner**
Computer Science Department
Stanford University
*uri@cs.stanford.edu*

**Ronald Parr**
Computer Science Department
Duke University
*parr@cs.duke.edu*

## Abstract

An important subclass of hybrid Bayesian networks are those that represent *Conditional Linear Gaussian (CLG)* distributions — a distribution with a multivariate Gaussian component for each instantiation of the discrete variables. In this paper we explore the problem of inference in CLGs, and provide complexity results for an important class of CLGs, which includes Switching Kalman Filters. In particular, we prove that even if the CLG is restricted to an extremely simple structure of a polytree, the inference task is NP-hard. Furthermore, we show that, unless P=NP, even approximate inference on these simple networks is intractable.

Given the often prohibitive computational cost of even approximate inference, we must take advantage of special domain properties which may enable efficient inference. We concentrate on the fault diagnosis domain, and explore several approximate inference algorithms. These algorithms try to find a small subset of Gaussians which are a good approximation to the full mixture distribution. We consider two Monte Carlo approaches and a novel approach that enumerates mixture components in order of prior probability. We compare these methods on a variety of problems and show that our novel algorithm is very promising for large, hybrid diagnosis problems.

## 1 Introduction

Bayesian networks are a useful modeling language for complex stochastic domains. Lately there has been a growing interest in *hybrid* models, which contain both discrete and continuous variables. An important class of hybrid models is *conditional linear Gaussian (CLG)* Bayesian networks. In these models, the conditional distribution of the continuous variables given the discrete ones is a multivariate Gaussian. CLG models are popular in a variety of applications, in both static and dynamic settings. Example applications include target tracking [1], where the continuous variables represent the state of one or more targets and the discrete variables might model the maneuver type; visual tracking, (e.g., [13]) where the continuous variables represent the head, legs, and torso position of a person and the discrete variables the type of movement; fault diagnosis [10], where discrete events can affect a continuous process; and speech recognition [7, ch.9], where a discrete phoneme determines a distribution over the acoustic signal.

The first part of the paper deals with the complexity of inference in CLG models. Although CLG models are commonly used, surprisingly little formal work has been done on analyzing their complexity. For the purposes of our analysis, we assume we are working with finite precision continuous variables and need to answer queries involving discrete variables. Thus, we are interested in solving questions of the form, "given some evidence $E$, what is the probability distribution of some discrete variable $A$?", or phrased as a decision problem, "given some evidence $E$, is the probability that a discrete variable $D$ takes on a the value $d$ in the range $(l, h)$?"

Obviously CLGs are a generalization of discrete Bayesian Networks, and therefore must be at least as difficult. However, it is not obvious whether network structures which are easy in the discrete case remain easy for CLGs. We prove this is *not* the case: Even for network structures for which inference in the discrete case is easy, inference for CLGs can still be NP-hard. In particular, we consider a very restricted class of CLG models, where the network structure is a polytree and every continuous variable has at most one (binary) discrete ancestor, and prove that even in this extraordinarily simple case, inference is NP-hard. After establishing that exact inference is NP-hard for these simple networks, we consider the question of approximate inference. We prove that unless P=NP there does not exist a polynomial time approximate inference algorithm with absolute error smaller than $0.5$. The class of networks we consider include Switching Kalman Filters [1, 6] as a special case; thus, we provide the first formal complexity results for this important class of models.

The second part of the paper addresses the question of how to perform inference in CLG models in light of our complexity results. The commonly used approach for CLG models is the algorithm proposed by Lauritzen [8, 9], which is an extension of the standard *clique tree* algorithm. Not surprisingly, since inference in CLGs is NP-hard even for simple networks, the size of the resulting clique tree is often exponential, leading to unacceptable performance.

In many domains, it is reasonable to expect that even though an exact answer might require an exponential number of Gaussians (possibly one for every assignment of the discrete variables), a small subset of these Gaussians is sufficient to produce a reasonable approximation. As we shall discuss, we believe that this is the case in a surprisingly large number of applications. Of course, the difficult part is efficiently generating this relatively small set. We consider two Monte Carlo approaches: stochastic sampling with likelihood weighting and MCMC. We also consider a novel approach that generates the Gaussians in order of their prior likelihood. We discuss the advantages and disadvantages of these three approaches and present some empirical results on a synthetic Bayesian Network.

To test our algorithm on a real-life domain we apply our techniques to the problem of tracking in *Dynamic Bayesian Networks* in the fault diagnosis domain. Here we need to track the state of the system where some discrete events (e.g., whether some fault happened) are hidden. The classical algorithms for this problem assume that there is only a small number of possible discrete events at every time step, and therefore do not scale up well for large systems. We show how our techniques, in combination with techniques described in [10], can circumvent this difficulty, leading to a practical algorithm for complex hybrid dynamic systems.

## 2 Preliminaries

*Bayesian networks (BNs)* are a compact graphical representation of probability distributions. Let $X_1, \ldots, X_n$ be a set of random variables, each of which takes values in some domain $\text{Dom}(X_i)$. A graphical model over $X_1, \ldots, X_n$ consists of two components: a directed acyclic graph whose nodes correspond to the random variables $X_1, \ldots, X_n$, and a set of *conditional probability distributions (CPDs)* $P(X_i \mid \text{Parents}(X_i))$. The structure of the network encodes a set of conditional independence assumptions that, together with the CPDs, uniquely define a joint distribution over $X_1, \ldots, X_n$. We use $\Delta$ to denote the discrete variables in the model and $\Gamma$ to denote the continuous ones.

The semantics allows for any type of CPD, involving both discrete and continuous variables. One particularly important subclass of these hybrid Bayesian networks are *conditional linear Gaussian (CLG)* models. In a CLG model, a discrete node cannot have continuous parents. Furthermore, the CPD of a continuous node is a conditional linear Gaussian, i.e., for every combination of the discrete parents the node is a weighted linear sum of its continuous parents with some Gaussian noise. More formally, if a node $X$ has continuous parents $\{Y_1, \ldots, Y_k\}$ and discrete parents $\boldsymbol{D} = \{D_1, \ldots, D_l\}$, we define its CPD using the following parameters: for every $\boldsymbol{d} \in \text{Dom}(\boldsymbol{D})$, we have $a_{\boldsymbol{d},0}, \ldots, a_{\boldsymbol{d},k}$ and $\sigma_{\boldsymbol{d}}^2$. The CPD is the defined as:

$$P(X \mid \boldsymbol{y}, \boldsymbol{d}) = Normal(a_{\boldsymbol{d},0} + \sum_{i=1}^{k} a_{\boldsymbol{d},i} y_i; \sigma_{\boldsymbol{d}}^2).$$

It can be easily shown that any CLG model represents a joint distribution with one multivariate Gaussian over the continuous variables for every instantiation of discrete variables. Conversely, any such distribution can be represented as a CLG model. We call each one of the possible discrete instantiations and the resulting Gaussian a *hypothesis*. Note that, in general, the number of hypotheses is exponential in the number of discrete variables.

## 3 NP-hardness of simple CLGs

The simplest class of discrete Bayesian Networks are polytrees, in which inference can be done in linear time. Therefore, it is important to ask whether we can perform inference efficiently in polytree CLGs. In this section we prove that unless $P = NP$ the answer is that we cannot. As stated before, we consider queries over some discrete variable given some evidence. Our goal is to analyze polytree CLGs where every continuous variable has at most one discrete ancestor, but we start with a simpler case:

**Theorem 1** *Inference in CLG models with binary discrete variables and a polytree graphical structure is NP-hard. Furthermore, unless P=NP there does not exist any polynomial approximate inference algorithm with absolute error smaller than* 0.5.

**Proof:** Consider the NP-complete *Subset Sum* problem [5]. We are given a set $S = \{s_1, s_2, \ldots, s_n\}$, where each element $s_i \in S$ is a non-negative integer, and a positive integer $L$. The question is whether there exists a subset $S' \subseteq S$ such that the sum of elements in $S'$ is exactly $L$.

We reduce this problem to a polytree CLG model, shown in Fig. 1(a). The discrete variables (shown as squares) are binary over $\{0, 1\}$ and have a uniform prior distribution. The CPDs of the continuous variables are:

$$P(X_1) = \begin{cases} Normal(0, \sigma^2) & A_1 = 0 \\ Normal(s_1, \sigma^2) & A_1 = 1 \end{cases}$$

For $i = 2, 3, \ldots, n$:

$$P(X_i) = \begin{cases} Normal(X_{i-1}, \sigma^2) & A_i = 0 \\ Normal(X_{i-1} + s_i, \sigma^2) & A_i = 1 \end{cases}$$

$$P(Y) = \begin{cases} Normal(L - \sqrt{2n}, 1) & B = 0 \\ Normal(X_n, \sigma^2) & B = 1 \end{cases}$$

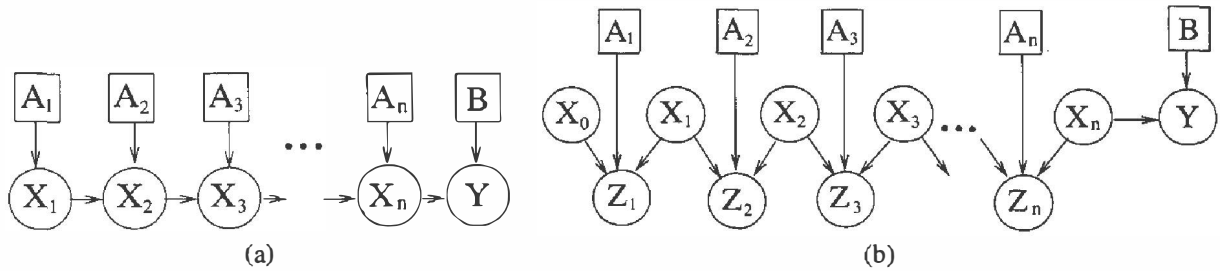

Figure 1: Networks used for the NP-hardness reductions

We choose $\sigma = \frac{1}{\sqrt{2Cn(n+1)}}$. $C$ is a constant which we will later use, but for now we can simply assume $C = 2$. We now prove that there exists a subset $S'$ whose elements sum up to $L$ iff $P(B = 1|Y = L) > 0.5$. Since $P(B)$ is uniform $P(B = 1|Y = L) > 0.5$ iff:

$$\frac{P(B=1|Y=L)}{P(B=0|Y=L)} = \frac{P(Y=L|B=1)}{P(Y=L|B=0)} > 1$$

First, we compute $P(Y = L|B = 0)$:

$$P(Y = L|B = 0) =$$

$$\frac{1}{\sqrt{2\pi}} \exp\left(-\frac{(L - \sqrt{2n} - L)^2}{2}\right) = \frac{1}{\sqrt{2\pi}e^n}$$

Let $a^k = \langle a_1, a_2, ..., a_k \rangle$ be some assignment to $\langle A_1, A_2, ..., A_k \rangle$ ($a_i \in \{0, 1\}$). Also, let $S(a^k) = \sum_{i=1}^{k} a_i s_i$. It is easy to show that for $1 \leq i \leq n$:

$$P(X_k | a^k) = Normal(S(a^k), k\sigma^2)$$

For the first direction, we assume that $\exists S' \subseteq S$ whose elements sum to $L$, and show that $P(B = 1|Y = L) > 0.5$. Define $a^n$ as $a_i = 1$ iff $s_i \in S'$. Clearly $S(a^n) = L$, and $P(Y = L|B = 1, a^n) = Normal(L, (n+1)\sigma^2)$.

$$P(Y = L|B = 1) \geq P(Y = L, a^n | B = 1) =$$

$$P(a^n|B = 1)P(Y = L|B = 1, a^n) =$$

$$\frac{1}{2^n} \cdot \frac{1}{\sqrt{2\pi(n+1)}\sigma} = \frac{\sqrt{2Cn}}{2^n \sqrt{2\pi}}$$

Therefore:

$$\frac{P(Y=L|B=1)}{P(Y=L|B=0)} \geq \frac{e^n}{2^n}\sqrt{2Cn} > 1$$

Conversely, we assume that there does not exist such $S' \subseteq S$ and show that $P(B = 1|Y = L) < 0.5$. Since $S'$ does not exist, we get that $\forall a^n$ $S(a^n)$ is an integer different from $L$, therefore $|L - S(a^n)| \geq 1$.

$$P(Y = L|B = 1) =$$

$$\sum_{a^n} P(a^n|B = 1)P(Y = L|B = 1, a^n) =$$

$$\sum_{a^n} \frac{1}{2^n} \cdot \frac{1}{\sqrt{2\pi(n+1)}\sigma} \exp\left(-\frac{(L - S(a^n))^2}{2(n+1)\sigma^2}\right) \leq$$

$$2^n \frac{1}{2^n} \frac{\sqrt{2Cn}}{\sqrt{2\pi}} \exp\left(\frac{-1}{2(n+1)\sigma^2}\right) = \frac{\sqrt{2Cn}}{\sqrt{2\pi}} e^{-Cn}$$

Therefore:

$$\frac{P(Y=L|B=1)}{P(Y=L|B=0)} \leq \frac{\sqrt{2Cn}e^n}{e^{Cn}} = \frac{\sqrt{2Cn}}{e^{(C-1)n}} < 1$$

We can now explain the use of the constant $C$. By making $C$ big enough, $P(B = 1|Y = L)$ gets arbitrarily close to 1 if $S'$ exists and arbitrarily close to 0 if it does not. We could then use an approximation algorithm with an absolute error of $\epsilon < 0.5$ to answer the decision problem: Construct a problem instance where $P(B = 1|Y = L) \leq \frac{0.5-\epsilon}{2}$ or $P(B = 1|Y = L) \geq \frac{0.5+\epsilon}{2}$ and answer "yes" iff the algorithm answers that $P(B = 1|Y = L) \geq 0.5$. Therefore, unless P=NP, there does not exist a polynomial time approximate inference algorithm with an absolute error smaller than 0.5. ∎

**Theorem 2** *Inference CLG models with binary discrete variables and a polytree graphical structure is NP-hard even if every continuous variable has at most one discrete ancestor. Furthermore, unless P=NP there does not exist any polynomial approximate inference algorithm with absolute error smaller than* $0.5$.

**Proof:** The reduction is very similar to the previous proof, but we have to use a different network structure. We use the structure shown in Fig. 1(b). Again, all the discrete variables are binary with uniform distribution, and:

$$P(X_0) = Normal(0, \sigma^2)$$

For $1 \leq i \leq n$:

$$P(X_i) = Normal(M, \sigma_2^2)$$

$$P(Z_i) = \begin{cases} Normal(X_i - X_{i-1}, \sigma_1^2) & A_i = 0 \\ Normal(X_i - X_{i-1} - s_i, \sigma_1^2) & A_i = 1 \end{cases}$$

$$P(Y) = \begin{cases} Normal(L - \sqrt{2n}, 1) & B = 0 \\ Normal(X_n, \sigma^2) & B = 1 \end{cases}$$

Where $M = \sum_j s_j$. Our query is $P(B|\mathbf{Z}^n = \mathbf{0}^n, Y = L)$, where $\mathbf{Z}^k = \mathbf{0}^k$ is a notation for $Z_1 = 0, ..., Z_k = 0$.

We now show how to choose the $\sigma$'s. Intuitively, $\sigma_2$ should be very big, representing a very weak prior on the $X_k$'s, while $\sigma_1$ should be very small. Then, with the evidence $Z_k = 0$, $X_k$ will have a distribution which is very close to $Normal(X_{k-1}, k\sigma^2)$ or $Normal(X_{k-1} + s_k, k\sigma^2)$, depending on the value of $A_i$. If we could get exactly this distribution, we would be in the same situation as in Theorem 1 and we could use the same proof. Although we cannot get exactly to the desired distribution we can get very close. Namely, we can get exactly the desired variance and be within $\epsilon$ away from the mean. For $\epsilon > 0$ we choose:

$$\sigma_1^2 = \frac{\epsilon}{n \cdot M} \; ; \; \sigma_2^2 = \sigma^2 \left(1 + \frac{\sigma^2}{\sigma_1^2}\right)$$

By induction we can prove that $\forall 1 \leq k \leq n$

$$P(X_k|a^k) = Normal(\mu_k, k\sigma^2)$$

Where $|\mu_k - S(a^k)| \leq \frac{k\epsilon}{n}$. We are now almost in the situation of Theorem 1. Instead of $P(X_n|a^n)$ having the distribution $Normal(S(a^n), n\sigma^2)$, it has the distribution $Normal(\mu_n, n\sigma^2)$, where $|\mu_n - S(a^n)| \leq \epsilon$. If we choose $\epsilon = \sigma$, we can easily modify the inequalities from Theorem 1 and prove the desired result. ∎

It is interesting to note that the fact that exact inference in polytree CLGs is NP-hard, is not very surprising by itself. Possibly the most popular CLG models are Switching Kalman Filters: See [1] for a good introduction to these models, and [6] for an up-to-date literature survey. If we unroll all the time steps of a Switching Kalman Filter, we get exactly the network in Fig. 1(a). The continuous variable at the end of the chain has all the discrete variables as ancestors, and therefore its distribution is a mixture of exponentially many Gaussians. It was therefore reasonable to assume that inference in this case would be intractable, although no formal proof was known. The results in this paper go beyond giving a formal proof for this intuition. First, we concentrate on queries involving just discrete variables whose posterior distribution is easy to represent (but not to infer). Second, in Theorem 2, the prior distribution of every continuous variable is either a Gaussian or a mixture of two Gaussians. Thus, it is somewhat more surprising that the seemingly benign structure of the continuous variables can produce an NP-hard decision problem. Perhaps even more important, and less obvious, is the fact that even the simplest type of approximate inference (an absolute error smaller than 0.5) is intractable.

**Corollary 3** *Finding the most likely instantiation of the discrete variables given some evidence is NP-hard.*

**Proof:** A direct result of the reductions: Had we known the most likely instantiation of the discretes, it would have been easy to determine whether $B$ was more likely to be 0 or 1 given the evidence. ∎

We conclude with a technical issue which is of interest mostly from a complexity theory point of view. Subset Sum is pseudo-polynomial; thus, it is possible that our NP-hardness results do not hold if all the parameters are polynomial in $n$ and $\frac{1}{n}$. To show that this is not the case, we can use the *Subset Product* problem [5, problem SP14] which is NP-complete in the strong sense. We are given a set $T = \{t_1, t_2, ..., t_n\}$ and a number $M$, where all the set elements $t_i \in T$ and $M$ are positive integers. The question is whether there exists a subset $T' \subseteq T$ such that the *product* of elements in $T'$ is exactly $M$.

The main idea is to use the same reduction as in Theorem 1. We set $s_i = \log(t_i)$ and $L = \log(M)$, and the question becomes whether we can find a subset of the $s_i$'s whose *sum* is exactly $L$ (note that these are not integers; thus, we do not have an instance of the Subset Sum problem). There are two technical issues that need to be addressed. First, the numbers $s_i$ and $L$ are not rational, and we must make sure that it is enough to represent them with precision that requires polynomial space. Second, the difference between the $s_i$'s is not fixed to 1 as before, and can become quite small. In particular, a subset of the $s_i$'s which does not sum up to $L$ can get very close to $L$ (unlike the situation before where the difference between $L$ and any subset which did not sum up to $L$ was at least 1). Fortunately, this difference is bounded from below by a polynomial in $\frac{1}{n}$, so we can modify the proof from Theorem 1 to work in this case, using means which are polynomials in $\log n$ and variances which are polynomials in $\frac{1}{n}$. The exact proof does not provide any further insights into CLGs; thus, we omit it from this paper.

## 4 Approximate Inference Algorithms

### 4.1 Domain Properties

Given that inference for very simple CLGs is NP-hard, and that even approximate inference is not tractable, one might conclude that inference in CLG models is a lost cause. Fortunately, many real life domains have special features which can be exploited by fast algorithms. In this paper we concentrate on the fault diagnosis domain. In this domain, it it is often the case that a relatively small subset of the mixture is a good approximation for the entire mixture. Thus, although we may need to deal with an exponential number of hypotheses to answer a query exactly, we can get good approximations if we can find a small subset which approximates the mixture well. Unfortunately, from Corrolary 3, even this problem is NP-hard.

Fortunately, the problem of finding the most likely hypotheses is simpler in the fault diagnosis domain. Since the probability of failures is relatively small, we know *a priori* that hypotheses which correspond to small numbers of faults are much more likely than the others. We note that this situation is not unique to the fault diagnosis domain. For example, in visual tracking based on past evidence many hypotheses are unlikely (e.g., given that two people are having a conversation, it is unlikely that one person would start running).

We consider two Monte Carlo approaches to this problem: sampling and MCMC. It turns out that sampling and MCMC both have difficulty with fault diagnosis because faults are generally quite unlikely and they are not sampled frequently. To overcome this difficulty, we consider a novel algorithm that uses the low probability of faults to its advantage: Our algorithm tries to find a good approximation for the full set of hypotheses by first considering the hypotheses which are more likely *a priori* (hypotheses with a small number of failures, or failures which tend to happen together). This idea of concentrating on likely hypotheses with a small number of faults is very natural, and was used in [3], although the probabilistic model there was discrete and some strong independence assumptions were made.

### 4.2 Monte Carlo Methods

We begin by presenting a naive algorithm that enumerates all the hypotheses resulting in a mixture of Gaussians. We show that sampling and, later, our incremental algorithm, can be viewed as approximations of this naive algorithm.

Formally, consider the general query of the form $P(Q_\Delta, Q_\Gamma \mid d, x)$ where $Q_\Delta, d$ are discrete and $Q_\Gamma, x$ are continuous. Instead of answering the query directly, it is easier to compute $P(Q_\Delta, Q_\Gamma, X \mid d)$. The result would be a mixture of Gaussians over $\{Q_\Gamma, X\}$ for every combination of $Q_\Delta$. We can then get the answer to the original query by conditioning every Gaussian in the mixture on the evidence $x$ and adjusting the weight of the mixture component accordingly (i.e., multiply the weight by the probability of $x$ given the multivariate Gaussian).

In order to compute $P(Q_\Delta, Q_\Gamma, X \mid d)$ we compute $P(q_\Delta, Q_\Gamma, X \mid d)$ for every possible value $q_\Delta$ of $Q_\Delta$. We do that by explicitly enumerating the possible instantiations $\delta$ of $\Delta$. Note that each instantiation $\delta$ defines single multivariate Gaussian distribution over $Q_\Gamma$, thus:

$$P(q_\Delta, Q_\Gamma, X \mid d) =$$
$$\sum_{\delta \in \mathrm{Dom}(\Delta)} P(\delta \mid d) P(q_\Delta, Q_\Gamma, X \mid \delta, d) =$$
$$\sum_{\delta \in \mathrm{Dom}(\Delta), q_\Delta \subseteq \delta} P(\delta \mid d) P(Q_\Gamma, X \mid \delta)$$

Where $q_\Delta \subseteq \delta$ means that $q_\Delta$ agrees with $\delta$ on the values of the variables they share. The resulting expression is easy to compute: $P(\delta \mid d)$ involves only discrete inference and $P(Q_\Gamma, X \mid \delta)$ is a multivariate Gaussian. The problem, of course, is the summation over all discrete combinations.

First, observe that we do not need to sum over all the discrete variables in the network. Instead, it is enough to sum over the discrete variables which determine the distribution of the continuous variables, i.e., the *direct parents* of the continuous nodes which we note as $\Delta_{\mathrm{dp}}$. Thus, from now on we assume that $\delta$ ranges only over $\Delta_1 \stackrel{\mathrm{def}}{=} Q_\Delta \cup \Delta_{\mathrm{dp}}$. We can still use the same summation, where we simply perform some different discrete inference for $P(\delta \mid d)$.

It is often the case that even summing just over $\Delta_1$ is infeasible. We present three methods that attempt to find a small subset of Gaussians which approximates the full sum. The first method is based on sampling — we use the probability distribution of $P(\Delta_1 \mid d)$ to sample assignments for $\Delta_1$, and then use these samples as our subset of hypotheses. To do so, we create a clique tree over the discrete variables, set the evidence $d$ and then sample from the tree, ignoring anything which is not in $\Delta_1$. One can view this method as a static version of *Rao Blackwellized Particle Filtering* or RBPF [4] — we sample the discrete variables and solve the remaining continuous problem analytically.

The sampling method runs into problems when the prior probability mass of a very small number of hypotheses dominates all the rest. This is typical in fault diagnosis problems, where for reasonably reliable systems, the prior probability of the "no fault" hypothesis can be bigger than 99%. In this case, we might waste most of our computational resources by generating duplicate samples instead of exploring other hypotheses. Indeed, for many systems this approach will be unlikely to ever generate a fault hypothesis with any reasonable number of samples. This means that our system will fail precisely when it is needed the most: when a fault has actually occurred.

The problem with this approach to sampling is that it generates hypotheses using $P(\Delta_1 \mid d)$, whereas ideally we wish to generate the hypotheses using their posterior distribution $P(\Delta_1 \mid d, x)$ (if we could do that, we would always generate the $K$ most likely hypotheses a-posteriori).

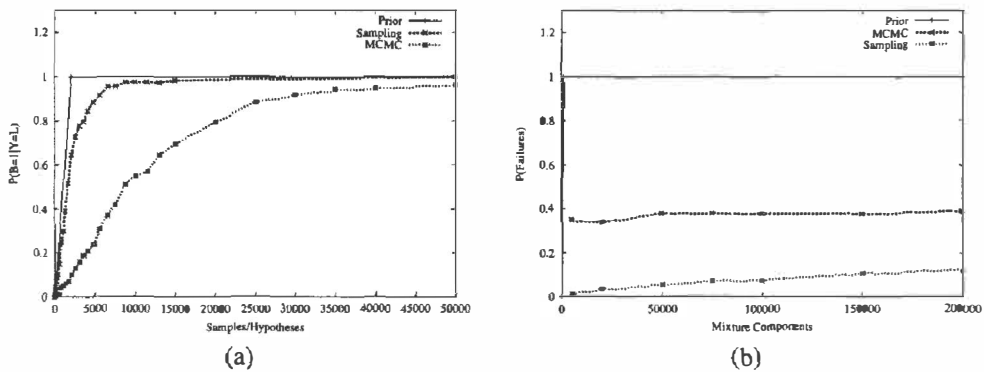

Figure 2: (a) Results on the Subset Sum Problem (b) Results on the unrolled five-tank network

An alternative approach is to try sampling hypotheses from the true posterior distribution, using an MCMC approach, namely Gibbs sampling [11]. To do so, we need to find the conditional distribution of some discrete variable $A_i$ given the rest of the variables in $\Delta_1$ and the continuous evidence. For some assignment $A_i = a$, let $\Delta_1[A_i = a]$ be the assignment $\Delta_1$ except that $A_i = a$. For every such $a$ we need to compute:

$$P(\Delta_1[A_i = a] \mid d, x) \propto P(\Delta_1[A_i = a], x|d) =$$

$$P(\Delta_1[A_i = a] \mid d)P(x|\Delta_1[A_i = a])$$

The first term involves a discrete inference and the second involves just one Gaussian, defined by $\Delta_1[A_i = a]$. It follows that the transition probabilities can be computed efficiently. Thus, we can use Gibbs sampling to generate samples from the correct posterior distribution $P(\Delta_1 \mid d, x)$, and then use these samples as our subset of hypotheses.

### 4.3 Generating in order of prior likelihood

A different approach to the problem of concentrated prior probability is to generate hypotheses deterministically in decreasing order of the probability $P(\Delta_1 \mid d)$, without any duplicates. With this method the algorithm uses all its run time to consider as many distinct hypotheses as possible. Note that we can ignore all the continuous variables in the BN for the purposes of the enumeration: we only care about the discrete problem of enumerating from $P(\Delta_1 \mid d)$.

The key subroutine of this method is an algorithm presented in [12] for enumerating the $K$ most likely configurations of a discrete BN given some evidence. The algorithm generates a clique tree and uses it to generate the configurations in an anytime fashion. The complexity of the algorithm is $O(|\Lambda| + Kn + Kc\log(Kc))$ where $|\Lambda|$ is the size of the clique tree (overall number of table entries), $c$ is the number of cliques, and $n$ is the number of variables.

The only difficulty in using this subroutine is the fact that we want to enumerate instantiations only of $\Delta_1$ rather than all the discrete variables. To do so, we need to create a clique tree just over $\Delta_1$. One way to do so is to run variable elimination without eliminating the variables from $\Delta_1$, and then build a clique tree from the remaining factors. Since this is done only once per network, and we assume the structure of the network is simple, this operation is relatively cheap. However, if the tree over $\Delta_1$ is too large, it is always possible to fall back and use the full clique tree to enumerate the full instantiations over $\Delta$.

### 4.4 Approximate inference discussion

We conclude with a few general comments. First, after we decide which hypotheses are in our representative subset, we need to determine their weights. One way to do so is to compute the discrete likelihood for each hypothesis, multiply by the continuous evidence likelihood and then normalize (ignoring duplicate hypotheses). Another way, possible for the sampling and MCMC based approaches, is to give weights based on the number of times each hypothesis was sampled multiplied by the continuous evidence likelihood. The latter is unbiased, but the former reduces the variance because it removes some of the randomness associated with sampling. Our experiments show that setting the weights using the likelihoods works well in practice.

An important issue is bounding the error of our approximation. One possible approach is to bound the probability mass of hypotheses which were not generated. This mass is the sum $\sum_\delta P(\delta \mid d)P(X \mid \delta)$ over $\delta$ which were not generated. To bound this sum, we must bound the densities $P(X \mid \delta)$ (which may be bigger than 1). It is easy to bound these densities if the network is a polytree and $X$ has just one variable by greedily minimizing the variance of every continuous variable in topological order. However, when the network is not a polytree or when $X$ has more than one variable, we must consider the covariances between variables, and the problem becomes more difficult: minimizing the variance of a single query variable over the discrete instantiations in general network structures is NP-hard. We hope to further address this problem in future work.

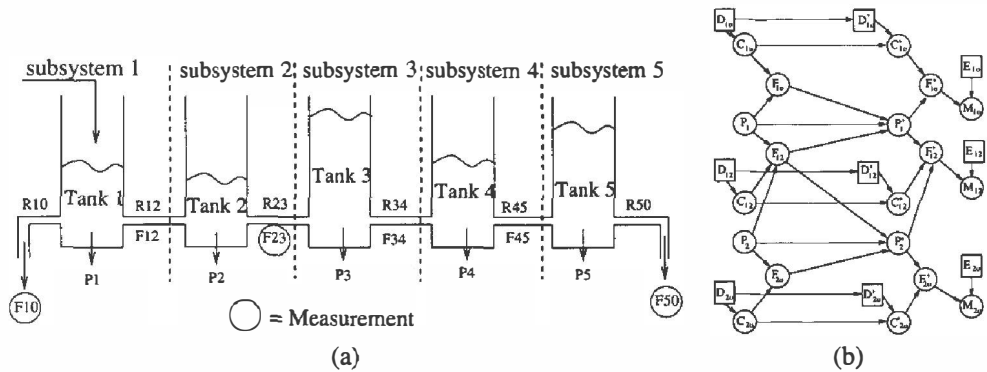

Figure 3: (a) The five tank system (b) DBN model (two tanks only)

So far, we assumed that the algorithm returns is a mixture of Gaussians for every instantiation of $Q_\Delta$. We can relax this assumption by letting the algorithm return only a single Gaussian for every instantiation of $Q_\Delta$, such that the Gaussian has the same first and second order moments as the mixture (similar to Lauritzen's algorithm). In this case, there is no need to keep all the Gaussians in memory — after generating a Gaussian and conditioning it on the evidence we can collapse it with the previous Gaussians and discard it, reducing the space complexity of our algorithm.

Another assumption was that we had to generate some hypotheses for every assignment to $Q_\Delta$. Alternatively, we can enumerate hypotheses globally, initially ignoring $Q_\Delta$, and then estimate the probability of these variables from the hypotheses we have generated. The former approach guarantees that we get some hypotheses for every assignment to the discretes in $q_\Delta$, while the latter may be more efficient if some of these assignments are extremely unlikely.

We conclude by comparing our algorithm with Lauritzen's algorithm. For simple networks our algorithm will enumerate all the hypotheses and will have the same complexity as Lauritzen's algorithm. In these cases, Lauritzen's algorithm is usually preferable, as it leads to more efficient implementation. However, for large networks our algorithm has two important advantages. First, it is an anytime algorithm — it can give some answer (albeit an approximate one) whenever we request it. The second advantage is space complexity: Lauritzen's algorithm performs *strong* triangulation, which forces all the direct parents to be in one clique, leading to exponentially sized cliques even for simple networks such as the networks in Fig. 1. In contrast, our storage requirements are dictated by the size of the query and are often exponentially smaller.

## 5 Results for static networks

We began by testing the algorithms on a network of the type used in Theorem 1. We considered the Subset Sum problem with 10 binary variables, picked a value for $L$ as evidence that corresponded to a valid subset sum, and then queried $P(B = 1|Y = L)$. The results are shown in Fig. 2(a). Note that the correct value is extremely close to 1. The Monte Carlo algorithms are averaged over 100 runs, while the plot for the enumeration algorithm was computed analytically. Recall that all discrete hypotheses have the same probability in this model; thus, the actual time at which the correct hypothesis is generated will be uniformly distributed over the number of hypotheses. We can therefore compute expectation in the number of samples required to find the correct hypothesis (we performed a few runs to confirm that the algorithm worked correctly). The main conclusion to draw from these experiments is that for uniformly likely discrete instantiations, enumeration is expensive but preferable to sampling. Note that we present the results as a function of the number of samples generated: all three algorithms spent much more time in generating the Gaussians corresponding to the discrete assignment than the discrete assignment itself; thus, we get a similar graph where the $X$ axis represents CPU time.

We next considered a variant on this model that is more closely related to the diagnosis domain. Instead of a uniform prior over discretes, we gave each discrete a probability of 0.999 and considered an $L$ that corresponded to two such events, i.e., a sum of two of the $s_i$. In the fault diagnosis domain, this would correspond to two simultaneous faults, a one in a million event. This may seem far-fetched, but it is not that unlikely in the lifetime of real system with a cycle time on the order of tenths of a second. We do not show a performance graph for this problem because neither sampling nor MCMC where able to find any reasonable hypotheses after 50000 samples. Enumeration from the prior found the correct hypothesis and concluded $P(B = 1|Y = L) = 1$ by considering up to 100 hypotheses. This is because the fault probabilities impose a partial ordering on hypotheses and all the "two failures hypotheses" are within the first 100 elements of this order.

Finally, we considered whether additional evidence would help the Monte Carlo methods catch up with the enumer-ation algorithm. We allowed sampling and MCMC to observe variable $X_5$ which, for the selected value of $L$, partitioned the problem in two halves, each containing a single subset element. This did not help sampling at all, but MCMC estimated $P(B = 1|Y = L) = 0.16$ after 50000 samples, a slight improvement. Enumeration from the prior does not take continuous evidence into account when generating hypotheses, so its performance was unaffected.

## 6 Application to Dynamic Systems

We hoped to evaluate these algorithms on a realistic, large, static CLG Bayesian network, but we had difficulty finding such a network (most likely because of the lack of efficient inference algorithms). Instead, we turned to dynamic systems, from which we can generate large and realistic static BNs. *Dynamic Bayesian Networks (DBNs)* are an extension of Bayesian networks to dynamic systems. In a DBN, time is partitioned into regular *time slices*. The state at each time slice is some value of the random variables $X^t$. The *transition model*, representing $P(X^{t+1} \mid X^t)$, and the *observation model*, representing $P(O^t \mid X^t)$, are represented using a BN fragment called a 2TBN. We can create large static Bayesian Networks from a DBN by considering the result of unrolling the network $k$ time steps. Dynamic systems are, perhaps, one of the most natural and important domains for CLG models, where they can be applied to such problems as visual tracking and fault diagnosis.

We evaluate the performance of sampling, MCMC and enumeration from the prior as applied to the diagnosis problem shown in Fig. 3(a). The physical system consists of five water tanks, connected by pipes, and includes measurements of the flows in three of the pipes. The continuous variables represent flows, pressures, conductances and flow measurements in the various pipes. The discrete variables represent the various failure modes. Fig. 3(b) shows the DBN model of the system (due to space considerations, the DBN describes a system with two water tanks rather than five). [1]

We unrolled this DBN by three time slices and created a scenario with two failures, a burst in the pipe between tanks 1 and 2, and a burst in the pipe between tanks 3 and 4. Fig. 2(b) shows the estimate of the probability that both pipes have burst given measurements over the three time slices. The correct value is very close to 1, but neither sampling nor MCMC found this after 200000 samples. Enumeration from the prior is guaranteed to find the double-failure after enumerating no more than 1089 hypotheses.

---

[1] An actual physical system would have a non-linear relation between pressures and flows, and cannot be described as a CLG. To deal with this problem we use a linear approximation, modeling the flow as the product of the pressure and the conductance. This approximation is appropriate for slow flows. We plan to use better approximations in the future, but for the purposes of testing our algorithm on simulated data, the physical accuracy of the model is not an issue, and one can treat it simply as a given CLG.

As a final test for our algorithm, we turn our attention to inference in actual DBNs, rather than unrolled static networks. The most common inference task in a DBN is to track the state of the system as it evolves, based on a sequence of observations $o^1, \ldots, o^t$. More precisely, the goal is to maintain a *belief state*, which is the distribution $P(X^t \mid o^1, \ldots, o^t)$. Once we have the belief state for time $t$ we can compute the new belief state for time $t + 1$ by performing inference in the 2TBN, treated as a static network.

If the system is large, keeping even one Gaussian for every possible combination of discrete variables in the belief state may be too expensive. Our algorithm can also be integrated effectively with the BK algorithm [2], which was adopted for hybrid systems in [10]. The key idea is to exploit the fact that large systems are often composed of subsystems, and while the subsystems are correlated, the interaction between them is often not so strong. The BK algorithm approximates the true belief state via separate belief states over the subsystems. We then plug these belief states into the DBN and use inference to find the belief states over time $t + 1$, accounting for the correlations between the subsystems. We can easily apply our inference algorithm to this task.

We modified the algorithm described in [10] to enumerate hypotheses from the prior distribution. In these systems it is interesting to track the state of the continuous variables over time, and these results are presented Fig. 4 (a). We picked the following very challenging sequence of events: $t = 5$: $C_{23}$ (conductance of the pipe between tanks 2 and 3) starts a negative drift; $t = 10$: simultaneous measurement failures of $F_{23}$ (flow between tanks 2 and 3) and $F_{5o}$ (flow out of tank 5); $t = 13$: $C_{23}$ bursts; $t = 17$, $C_{45}$ starts a negative drift; $t = 23$, $C_{45}$ bursts; $t = 25$, $C_{12}$ bursts. For this track, we show the belief state against ground truth for three continuous variables: $C_{12}$, $C_{45}$ and $P_5$ (the pressure in tank 5). We also considered the probability of discrete variables, which tracked the true events correctly (we do not show those due to space considerations).

It is interesting, although somewhat unfair, to compare our algorithm to an "omniscient" Kalman filter. The omniscient Kalman filter (Fig. 4(b)) knows the value of every discrete variable at every time step, and needs to track only the continuous state. Clearly, this should do a much better job of tracking the system than any method based on imperfect, partial information. Nevertheless, our algorithm mirrors this gold standard very closely, reflecting the success of our algorithm in finding the correct hypotheses.

## 7 Discussion and conclusions

We have proven that even simple CLG models can be very challenging. We provided the first NP-hardness results for a very simple class of CLG models, and also showed that unless P=NP there is no efficient approximate infer-ence algorithm for these cases. As an immediate corollary, we provided complexity results for the important class of Switching Kalman Filters. We stress that our results do not imply that CLGs are not useful, but rather suggest that in order to use them efficiently one must make some assumptions or observations under which inference or approximate inference becomes tractable (e.g., the small number of likely hypotheses in fault diagnosis representing a small number of faults or faults that tend to happen simultaneously). We feel that characterizing such sub-classes of CLGs is still an open problem, requiring further work.

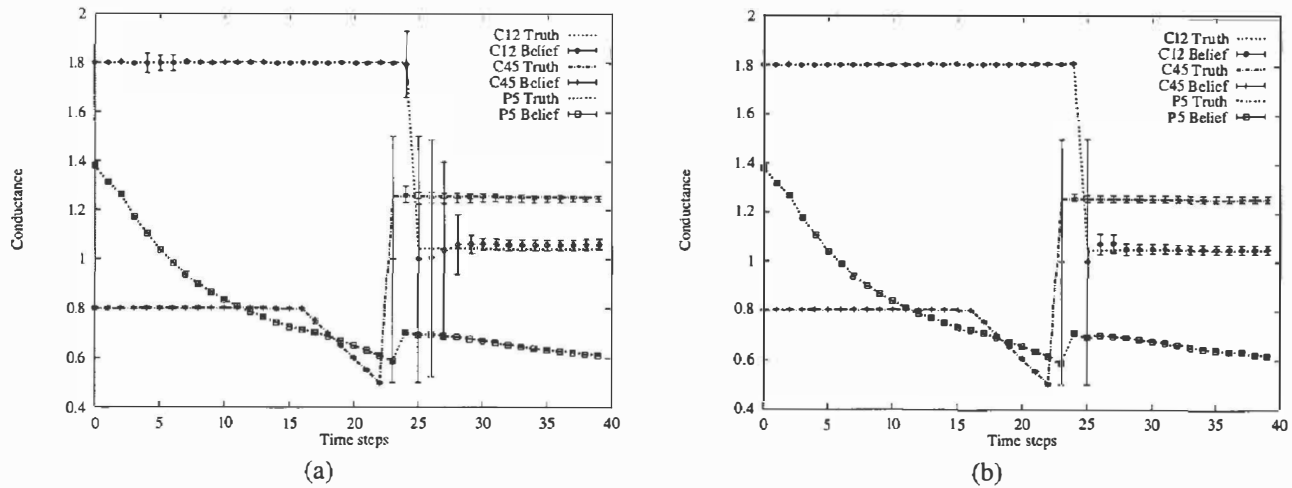

Figure 4: Two runs on a difficult trajectory for the five-tank system: (a) our algorithm; (b) an omniscient Kalman filter.

We compared several approximate inference algorithms, all of which have several advantages over exact inference. They are anytime in nature and have reduced space complexity. We presented a novel algorithm which takes advantage of the partial ordering on combinations of faults imposed by most fault diagnosis problems by generating hypotheses in decreasing order of likelihood. We found that the incremental algorithm has superior performance to sampling and MCMC for the type of unlikely evidence that is of greatest importance in fault diagnosis domains.

While inference for continuous variables was not an emphasis of this paper, the superior tracking of discrete variables also leads superior tracking of continuous variables. We are optimistic that the incremental algorithm for generating hypotheses, combined with the architecture presented in [10] will provide basis for efficient and robust tracking and diagnosis for many large systems of practical interest.

#### Acknowledgments.

We are pleased to thank Daphne Koller, Carlos Guestrin and Simon Tong for many useful discussions. This research was supported by ONR Young Investigator (PECASE) under grant number N00014-99-1-0464, and by ONR under the MURI program "Decision Making under Uncertainty", grant number N00014-00-1-0637.